\documentclass[10pt,twocolumn,letterpaper]{article}

\usepackage{cvpr}              

\usepackage{graphicx}
\usepackage{amsmath}
\usepackage{amssymb}
\usepackage{booktabs}

\usepackage{algorithm}
\usepackage{algpseudocode}
\usepackage{tabularx}
\usepackage{tabularray}
\usepackage{subcaption}

\usepackage[pagebackref,breaklinks,colorlinks]{hyperref}

\usepackage[capitalize]{cleveref}
\crefname{section}{Sec.}{Secs.}
\Crefname{section}{Section}{Sections}
\Crefname{table}{Table}{Tables}
\crefname{table}{Tab.}{Tabs.}

\def\cvprPaperID{*****} 
\def\confName{CVPR}
\def\confYear{2022}

\usepackage{etoolbox}
\usepackage{silence}
\makeatletter
\robustify\@latex@warning@no@line
\makeatother
\usepackage{authblk}

\title{Balanced 3DGS: Gaussian-wise Parallelism Rendering  with Fine-Grained Tiling}

\author{Hao Gui$^{1}$ \quad Lin Hu$^{2}$ \quad Rui Chen$^{2}$ \quad Mingxiao Huang$^{1}$ \quad Yuxin Yin$^1$ \quad Jin Yang$^{1}$ \\ Yong Wu$^{1}$ \quad Chen Liu$^{1}$ \quad Zhongxu Sun$^{1}$ \quad Xueyang Zhang$^{1}$ \quad Kun Zhan$^{1}$  \vspace{0.3em} \\
{\normalsize $^1$ Li Auto, Shanghai, China} \quad
{\normalsize $^2$ NVIDIA, Beijing, China}
}

\begin{document}
\maketitle
\begin{abstract}
   3D Gaussian Splatting (3DGS) is increasingly attracting attention in both academia and industry owing to its superior visual quality and rendering speed. However, training a 3DGS model remains a time-intensive task, especially in load imbalance scenarios where workload diversity among pixels and Gaussian spheres causes poor renderCUDA kernel performance. We introduce Balanced 3DGS, a Gaussian-wise parallelism rendering with fine-grained tiling approach in 3DGS training process, perfectly solving load-imbalance issues. First, we innovatively introduce the inter-block dynamic workload distribution technique to map workloads to Streaming Multiprocessor(SM) resources within a single GPU dynamically, which constitutes the foundation of load balancing. Second, we are the first to propose the Gaussian-wise parallel rendering technique to significantly reduce workload divergence inside a warp, which serves as a critical component in addressing load imbalance. Based on the above two methods, we further creatively put forward the fine-grained combined load balancing technique to uniformly distribute workload across all SMs, which boosts the forward renderCUDA kernel performance by up to 7.52x. Besides, we present a self-adaptive render kernel selection strategy during the 3DGS training process based on different load-balance situations, which effectively improves training efficiency.
\end{abstract}

\section{Introduction}
\label{sec:intro}
\noindent 3DGS \cite{3DGS-inria} is an innovative technique in 3D graphics that facilitates the reconstruction of detailed 3D scenes from images and viewpoints, allowing real-time multi-angle rendering. While it holds great promise, it also encounters significant computational challenges, which underscores the necessity for optimization \cite{chen2024survey,Survey01, Survey03, scalingup}. To boost 3DGS performance, two primary optimization strategies are currently being explored. The first strategy focuses on the modification of the algorithm itself to achieve better precision, fewer storage, or higher computational efficiency \cite{ali2024trimmingfatefficientcompression,HAC,GeoGaussian,3DGS-LM,CompGS,lee2024compact3dgaussianrepresentation,li2024dngaussianoptimizingsparseview3d,HexPlane,luiten2023dynamic3dgaussianstracking,3DGStream,PhysGaussian}. The second strategy, rather than making modifications to the algorithm itself, aims to enhance computational efficiency of raw algorithm by means of a variety of strategies, such as the implementation of intelligent task scheduling or the optimization of CUDA kernels \cite{DISTWAR,Taming, Mini-Splatting, gsplat, EAGLES, scalingup}.

Despite great advancements in 3DGS technology, training a 3DGS model remains a challenging and time-consuming task, particularly when faced with load imbalance situations. In such cases, imbalanced workload between pixels and Gaussian spheres can hinder training performance. Specifically, graphics processors (GPUs) are designed to handle regular and homogeneous tasks, and threads on GPUs run in SIMT (Single Instruction Multiple Threads) manner. Therefore, discrepancies in workload between pixels and Gaussian spheres can result in some threads being active while others remain idle, thereby degrading overall performance. To the best of our knowledge, we are the first to address those imbalance issues in kernel level. Through in-depth analysis and experimentation during the 3DGS training process, we have identified and characterized three significant load imbalance problems, which are as follows:\\\\
\textbf{Load imbalance on SMs due to CUDA's static distribution}: there is a load imbalance issue on SMs if we use static distribution. The renderCUDA dispatches tasks to SMs statically based on image and thread block size. Although the number of thread block tasks among different SMs is the same, the workload of each thread block can vary widely. This is a performance factor that has been neglected and results in a load imbalance between SMs.\\\\
\textbf{Load imbalance in tiles}: there is an issue of load imbalance among tiles. In 3DGS, the image is initially divided into multiple non-overlapping tiles to avoid the cost of deriving Gaussians for each pixel. Each tile contains 16×16 pixels as the existing method \cite{3DGS-inria}. However, some tiles may have extreme workloads compared with other light-workload tiles, if the basic CUDA distribution and tiling strategy remain unchanged, a severe load imbalance among SMs will occur.\\\\
\textbf{Load imbalance across training stages}: During the 3DGS training, the data characteristics change remarkably in different phases. At the initial stage, metrics such as Gaussian distribution have significant differences among various blocks or threads. As the training progresses, these imbalances get improved as data characteristics change. After numerous iterations, extreme data metrics become more balanced.

To address these issues comprehensively, we introduce "Balanced 3DGS". The highlights include: We are the first to propose the Gaussian-based parallel load balancing method to optimize the forward computation of the render CUDA kernel within warps. Additionally, we innovatively present inter-block dynamic workload distribution to boost the performance of the forward of render CUDA kernel. By evenly distributing tasks among computation blocks, it minimizes idle time of threads and maximizes resource utilization. Our fine-grained combined load balancing approach combines these two techniques, providing a complete solution to load imbalance problems. We propose an experiment-based adaptive kernel selection strategy. This strategy overcomes the limitations of manual metric selection, presents a more accurate reflection of training process, and ensures better performance and efficiency.

\section{Background}
\subsection{3DGS}
\noindent The 3D Gaussian can be mathematically expressed as:
\begin{equation}
  G(x)=e^{-\frac{1}{2} (x-\mu )^{T} {\textstyle \sum_{}^{-1}(x-\mu )}  } 
  \label{eq:1}
\end{equation}

where $x$  represents an arbitrary position within the 3D scene. Each 3D Gaussian splat is allocated a position(mean $\mu$ ), and $\sum$  represents the covariance matrix of the 3D Gaussian function, and it is a positive definite matrix.
 
The Gaussian splatting method utilizes splatting techniques to project the 3D Gaussians onto 2D image planes for the purpose of rendering. Given the viewing transformation  $W$  and the Jacobian of the affine approximation for the projective transformation   $J$  , the covariance matrix   $ {\textstyle \sum_{}^{'}} $  in camera coordinates can be computed as below:

\begin{equation}
 {\textstyle \sum_{}^{'}} =JW\sum W^{T} J^{T} 
  \label{eq:2}
\end{equation}

Then 3D Gaussian $G(x)$ is tranformed to 2D Gaussian $G'(x)$ as below:

\begin{equation}
  G'(x)=e^{-\frac{1}{2} (x-\mu )^{T} {\textstyle \sum_{}^{'-1}(x-\mu )}  } 
  \label{eq:3}
\end{equation}

Each 3D Gaussian is composed of its position, color represented by spherical harmonics, opacity, rotation, and scaling. For a given pixel, the color blending of N-sorted 2D Gaussians is determined by:

\begin{equation}
C=\sum_{i\in N}^{} c_{i} \alpha _{i}\prod_{j=1}^{i-1} (1-\alpha _{j} )
\label{eq:4}
\end{equation}

where  $c_{i} $  designates the learned color of a pixel and $\alpha _{i}$ designates the multiplication result of an opacity and its 2D Gaussian $G'(x)$.

\subsection{CUDA Related}

\begin{figure}[htbp]
  \centering
   \includegraphics[width=0.95\linewidth]{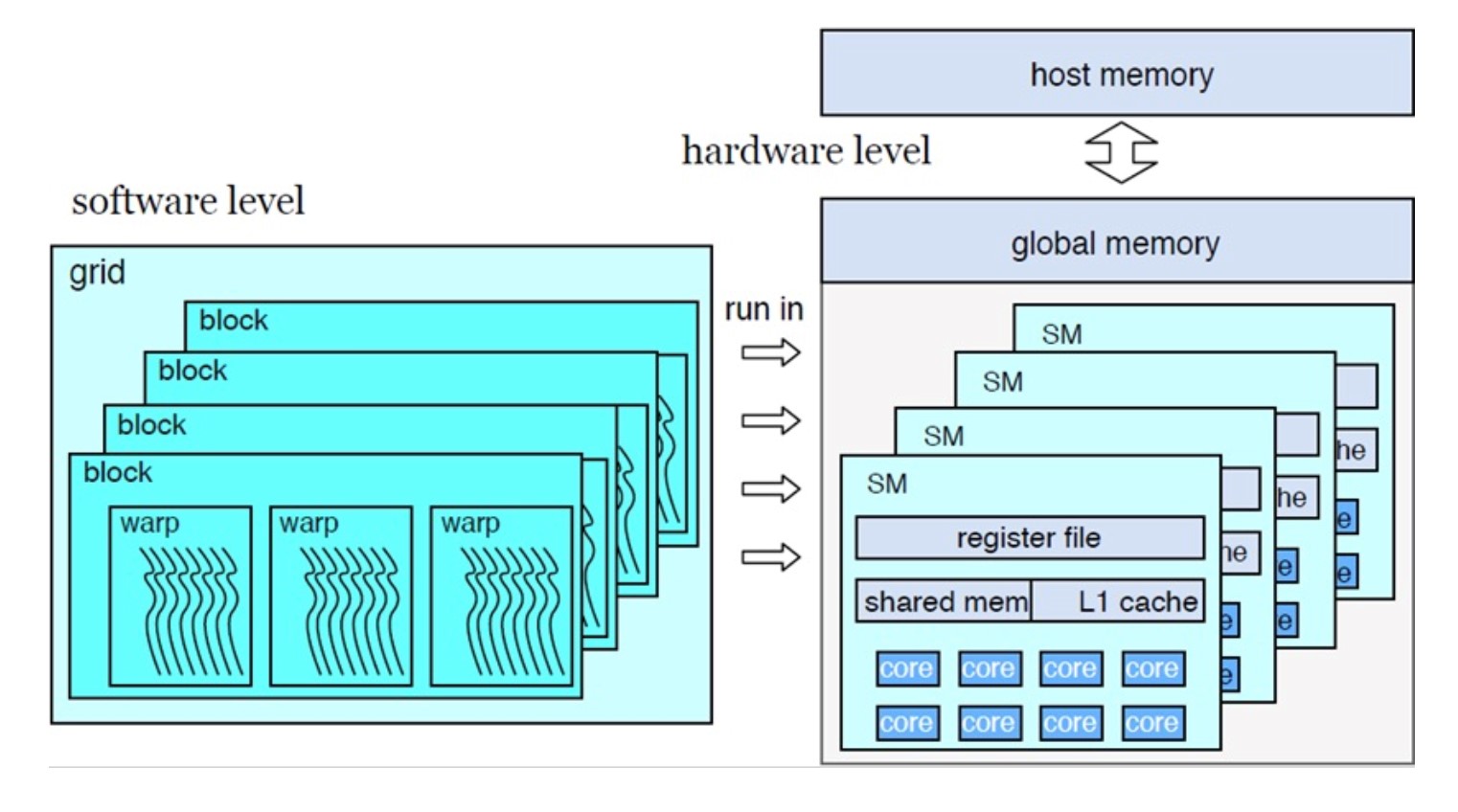}
   \caption{GPU Architecture.}
   \label{fig:gpuarch}
\end{figure}

\noindent This subsection presents an in-depth introduction to  GPU architecture and CUDA, a parallel computing platform and programming model developed by NVIDIA, including many key concepts that will be used in this paper such as thread, warp, block, shared memory and SM, as shown in Fig. \ref{fig:gpuarch}.

These concepts are fundamental to understanding how parallel computing is achieved and optimized on NVIDIA GPUs, enabling developers to harness the immense computational power of these devices for a wide range of applications including scientific computing, computer graphics, and machine learning.

\noindent \textbf{Thread}: a thread in the CUDA context is the basic unit of execution. It represents an independent sequence of instructions that can be executed in parallel with other threads. Each thread has its own program counter, register set, and local memory. Different threads can be used to process different tasks.

\noindent \textbf{Warp}: a warp is a key concept in the execution model of modern GPUs. A warp usually consists 32 threads, and those threads run in SIMT manner. This means that all the threads in a warp execute the same instruction simultaneously, but they may operate on different data. If threads within a warp diverge in their execution paths (due to conditional statements, for example), the GPU may serialize the execution of those diverging paths to ensure correct results, but this serialization will cause severe performance  drop. Thus, we should avoid thread divergence as much as we can.

\noindent \textbf{Block}: A block is a collection of threads that share a certain level of resources. The threads within a block have the ability to communicate and synchronize with each other through shared memory and synchronization primitives. Shared memory allows for fast data sharing among threads in the same block, which is beneficial for tasks that require data sharing and cooperation, such as parallel reduction operations. Multiple blocks are organized to a grid. 

\noindent \textbf{Shared Memory}: shared memory is a type of memory that is accessible to all threads within a block. It allows for fast communication and data sharing among the threads in a block. Shared memory is much faster than global memory, and it is common sense that proper use of shared memory can lead to significant performance improvements.

\noindent \textbf{SM}: SM is a major hardware component of the GPU architecture. It contains multiple CUDA cores, along with other resources such as register files, shared memory, and cache. When a CUDA program is launched, thread blocks are mapped to SMs, and the SMs then manage the execution of the threads within those blocks. Note that one block can only be mapped one SM, but one SM can have multiple blocks. The performance of the GPU is closely related to the efficiency of the SMs in handling workload.

\section{Related Works}

\noindent 3D Gaussian Splatting \cite{3DGS-inria} has emerged as one of the most prominent novel view synthesis (NVS) and rendering techniques \cite{eisemann2008floating, chaurasia2013depth, hedman2018deep, kopanas2021point} through the use of 3D Gaussian point cloud representations. It offers faster real-time rendering and training while maintaining high-quality output, outperforming previous NVS approaches such as NeRF \cite{NeRF}. Despite its advancements, the rendering efficiency of 3D Gaussian Splatting remains a critical area for improvement, particularly as the image scales increase. Ensuring efficient and high-quality rendering in such contexts remains a challenge \cite{Survey01, chen2024survey, Survey03}. Numerous efforts \cite{ma20243d, duan20244d, radl2024stopthepop, yu2024mip, peng2024rtg, ye20243d} have focused on algorithmic optimizations, while others have explored engineering solutions to enhance performance.\\

\noindent \textbf{Improving 3DGS Algorithms}: Traditional 3D Gaussian Splatting combines the primitive-based representations and volumetric representations, but this approach often results in reliance on redundant Gaussians, leading to overfitting on training views and insufficient handling of texture deterioration \cite{cheng2021multiview3dreconstructiontextureless}, lighting effects \cite{gortler2023lumigraph, levoy2023light, buehler2001unstructured} and the underlying geometric structures. To address these challenges, Scaffold-GS \cite{Scaffold-GS} introduces anchor points from Structure from Motion (SfM) \cite{SfM} to capture the geometric structure and optimize end-to-end training reconstruction. Similarly, the method in \cite{ali2024trimmingfatefficientcompression} employs a pruning strategy using a predetermined opacity threshold to remove redundant Gaussians, thereby accelerating computations. Building on Scaffold-GS, HAC \cite{HAC} implements a binary hash grid to establish the inherent spatial relationships among contiguous anchor points, achieving greater size reduction than \cite{Scaffold-GS}. However, SfM technology often struggles with producing adequate point clouds on texture-less surfaces, which can negatively impact rendering quality and alignment. In contrast to traditional 3DGS using basic split and copy strategy, GaussianPro \cite{GaussianPro} provides a progressive propagation strategy, leveraging the prior existing geometry and multi-view stereo (MVS) technique \cite{Jensen_2014_CVPR, goesele2007multi} to generate new spatial-oriented Gaussians for these challenging surfaces. GeoGaussian \cite{GeoGaussian} also applies mathematical smoothness constraints to improve alignment in the textureless regions such as ceilings, furniture and walls. Additionally, \cite{3DGS-LM} develops a tailored Levenberg-Marquardt (LM) replacing the traditional ADAM optimizer, further enhancing performance. CompGS \cite{CompGS} introduces vector quantization techniques such as K-means, significantly reduing storage and rendering demands with minimal quality loss. Following this line of work, other techniques \cite{lee2024compact3dgaussianrepresentation, yan2024multiscale3dgaussiansplatting, zhang2024fregs3dgaussiansplatting, li2024dngaussianoptimizingsparseview3d} have been developed to refine 3DGS through strategies like learnable masks, multi-scale approaches, frequently regularization, and depth regularization, leading to improved fidelity and efficiency in dynamic scene synthesis. Technological improvements also extend to dynamic environments, with methods such as HexPlane \cite{HexPlane} reducing training times through efficient representation using learned planar features while maintaining competitive view synthesis quality. Moreover, works like Dynamic 3D Gaussians \cite{luiten2023dynamic3dgaussianstracking} and 3DGStream \cite{3DGStream} advance dynamic scene rendering and tracking, highlighting innovations in first-person view synthesis and 4D video editing via real-time Gaussian transformations. PhysGaussian \cite{PhysGaussian} further enhances Gaussians with Newtonian dynamics, facilitating high-quality motion synthesis without traditional geometric constructs.\\

\noindent \textbf{Accelerating 3DGS Training}: Most contemporary approaches expedite training by either decreasing the number of Gaussians or improving the rasterizer. However, the exploration of training optimization within this research context remains limited. Typically, 3DGS models are trained on a single GPU and are often constrained by the memory bound \cite{Mvsplat, papantonakis2024reducing, HAC, niedermayr2024compressed} or the compute bound. We found that most common examples of training optimization are built on backward-render kernels \cite{DISTWAR}, with inadequate attention given to forward-kernel improvements. This imbalance highlights forward-render kernel optimization as a critical bottleneck for attaining efficient end-to-end 3D Gaussian Splatting training, which forms the primary motivation for our work. In 3DGS backward propagation, atomic operations on the L2 cache cause bottlenecks during gradient computation. DISTWAR \cite{DISTWAR} addresses this by dynamically distributing these operations between SM warp and L2 atomic partitions, reducing L2 cache load and atomicAdd calls, which improves GPU throughput and training efficiency. Mini-Splatting \cite{Mini-Splatting} observes that the inefficient spatial locality of 3D Gaussian point clouds representation where only a small number of Gaussians affect each image tile, contributing equally to the limitation of the model performance. To address this, \cite{Taming, Mini-Splatting} introduce the densification algorithm which reduces the unnecessary model size and \cite{Taming} also shifts parallelism from a traditional per-pixel to a per-splat approach for efficient backpropagation. FlashGS \cite{FlashGS} employs an exact intersection algorithm to remove overlapped unnecessary Gaussians on the preprocessing stage, while gsplat \cite{gsplat, EAGLES} supports a CUDA-accelerated differentiable rasterization library for 3DGS. Grendel \cite{scalingup} addresses the computational challenges that arise from dynamic variation in the distribution, shape, color, opacity, and density of Gaussians throughout the training process by implementing a distributed system involving multiple GPUs \cite{RetinaGS, Hierarchical-GS, DOGS, City-on-web}. By using mixed parallelism and spatial locality of 3DGS, Grendel tackles the issue of dynamic load imbalance. However, we focus on the load imbalance within SMs of the individual GPU and implement Gaussian-wise parallelism at the CUDA rendering level.

\section{Approach}

\noindent In order to address the load imbalance issues across SMs, tiles and training stages  mentioned above (in Section \ref{sec:intro}), we introduce Balanced 3DGS, a Gaussian-wise parallelism rendering with fine-grained tiling approach on 3DGS training. The key of this approach is how to schedule the workloads among different SMs considering the skewed distribution of Gaussians among different image tiles, so that we can achieve an optimal utilization in Gaussian rendering. Besides, it is worth noting that this is a solution without any precision loss.

\subsection{Inter-Block Dynamic Workload Distribution}
\label{sect:4.1}
\noindent If image tiles are mapped to thread blocks (SMs) in a static method, severe imbalance in Gaussian numbers in different image tile lead to severe load imbalance: some thread blocks will finish their jobs early, while we have to wait until the last thread block with the heaviest workload to finish, and this causes obvious overall performance drop and great compute resources waste. 

\begin{figure}[htbp]
  \centering
   \includegraphics[width=0.95\linewidth]{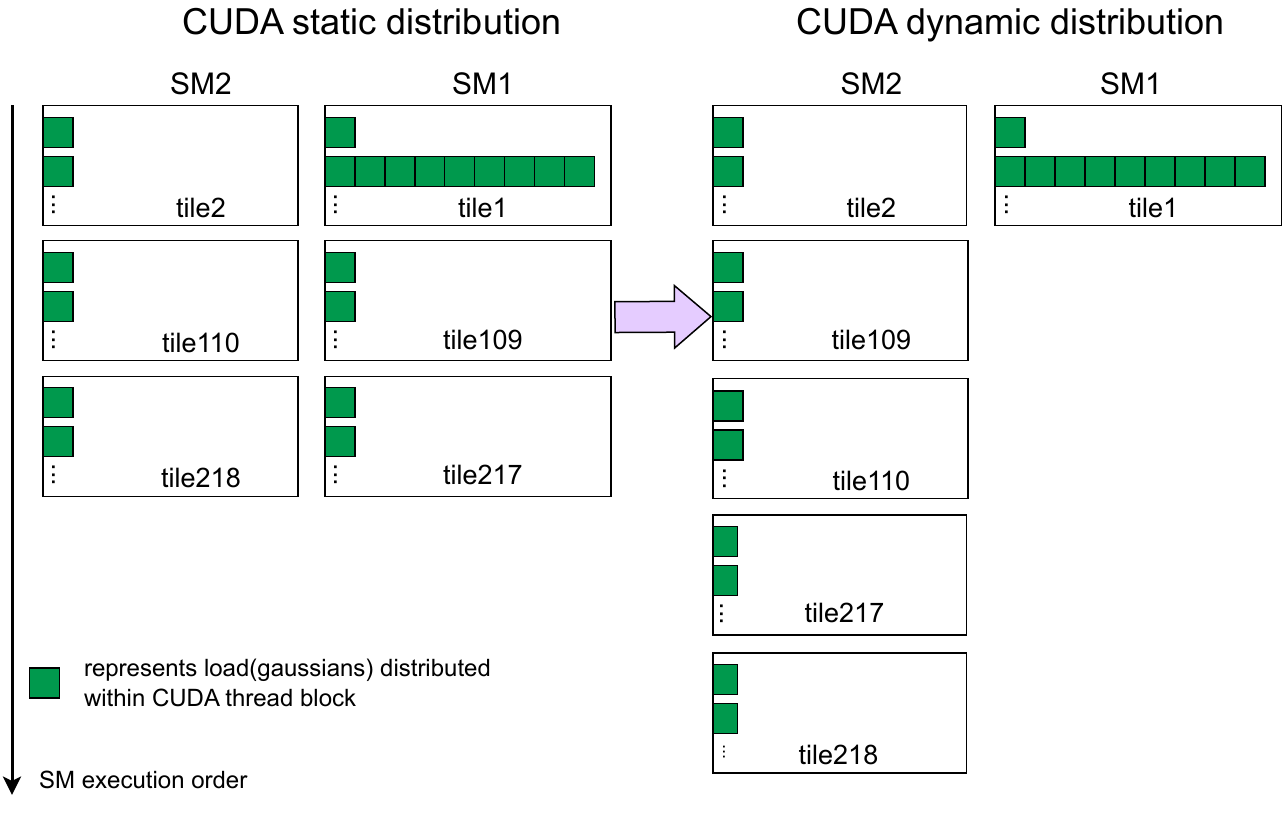}
   \caption{CUDA SM load distribution diagram.}
   \label{fig:cudasmload}
\end{figure}

To solve this, we use a dynamic task mapping (distribution) method: 1) tasks of all tiles are collected into a task pool; 2) all thread blocks will get their initial task first; 3) once a thread block finishes its current work, it will immediately fetch next tile from the task pool and run it; 4) all tasks are fetched by thread blocks in such a dynamic method(Algorithm \ref{alg:inter-block-load-balancing}). In this way, some thread blocks can focus on a small number of (mostly one) tile(s) with large workloads, while some blocks can deal with many tiles with small workload, thus avoiding idle SMs due to imbalanced workload distribution. Fig. \ref{fig:cudasmload} shows an example of SM workload difference between static and dynamic workload distribution.

\begin{algorithm}
\caption{Inter-Block Dynamic Load Balancing Render Computations}
\label{alg:inter-block-load-balancing}
\algrenewcommand\algorithmicrequire{\textbf{Input:}}
\algrenewcommand\algorithmicensure{\textbf{Output:}}
\footnotesize
\begin{algorithmic}[1] 
\Require point\_xy\_image(XY), point\_list(PL), conic\_opacity(CO), features(F), depths(D)
\Ensure out\_alpha(OA), out\_color(OC), out\_depth(OD)
\Function{dynamicAllocateRenderCUDA}{total\_tiles, next\_tile}
    \While {true}
        \State Synchronize block
        \If {thread\_idx $==$ 0}
            \State my\_tile\_id $\leftarrow$ atomicAdd(next\_tile, 1)
        \EndIf
        \State Synchronize block
        \If {my\_tile\_id $>$ total\_tiles}
            \State return
        \EndIf
        \State Initialize local variables c(color), w(weight), d(depth), t
        \State Load XY, PL, CO to share memory
        \For {j in all gaussians}
            \State BlendInOrder(XY,PL,CO,F,D,c,w,d,t,j) $\hspace*{\fill}$ $\triangleright$ Pixel Wise Parallel
        \EndFor
        \State OA, OC, OD $\leftarrow$ WriteOutputs(c, w, d, t)
    \EndWhile
\EndFunction
\\
\Function{kernelLaunch}{}
    \State total\_tiles $\leftarrow$ grid.x $\times$ grid.y $\times$ grid.z, each block deal with 128 pixels
    \State next\_tile $\leftarrow$ 0
    \State max\_hw\_resource $\leftarrow$ 108 (A100 SM) $\times$ 16 (16 blocks with 4 warps in a block)
    \State dynamicAllocateRenderCUDA$<<<max\_hw\_resource, block>>>$(total\_tiles, next\_tile)
\EndFunction

\end{algorithmic}
\end{algorithm}

\subsection{Gaussian-Wise Parallel Rendering}
\label{sect:4.2}
\noindent Currently, one thread is responsible for processing workloads of one pixel. And the naive render kernel uses pixel-level parallelism, while processes Gaussians in serial. But even inside an image tile where pixels share the same Gaussian group, the actual workloads of different pixels may vary a lot. Because if some of the pixels inside a tile finish their render process in a specific Gaussian, they don't have to iterate over the remaining ones. We call this "early stop". Early stops may cause large variance among workloads of pixels inside a tile. Thus, simply mapping pixels to threads will lead to load imbalance among threads inside a warp. Considering that threads inside a warp run in a SIMT manner, this imbalance also causes a great performance drop. 

Accordingly, we propose a warp-collaboration method. All threads inside the same warp will collaborate to process one single pixel by doing Gaussian-wise parallelism to avoid workload divergence inside a warp. In this method, the 32 pixels processed in parallel by a warp in the naive method will be processed in serial. In conclusion, we replace pixel-wise parallelism with Gaussian-wise parallelism to reduce workload divergence inside a warp (Algorithm \ref{alg:gaussian-wise-parallel-render}).

\begin{figure}[htbp]
  \centering
   \includegraphics[width=0.95\linewidth]{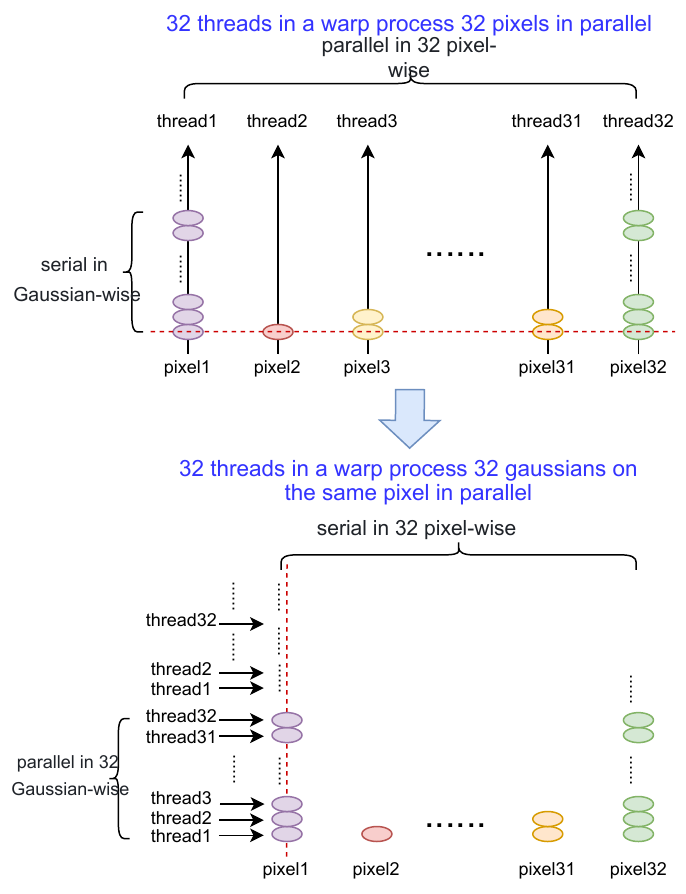}
   \caption{Comparison between pixel-wise parallelism and Gaussian-wise parallelism.}
   \label{fig:comparisons}
\end{figure}

In fact, there is dependency in the calculation of different Gaussians of a pixel, as revealed in above Equation \ref{eq:4}. Since we process Gaussians in 32-way parallelism, we need to do one more prefix multiplication operation to assure each Gaussian obtains all results from its precursors inside the 32-size Gaussian group \footnote{Detailed Implementation See Algorithm \ref{alg:detail gaussian-wise parallel render} line 12 to line 17.}. Fig. \ref{fig:comparisons} shows the difference in warp-level pixel-wise parallelism and warp-level Gaussian-wise parallelism.

\begin{algorithm}
\caption{Gaussian-Wise Parallel Render Computations}
\label{alg:gaussian-wise-parallel-render}
\algrenewcommand\algorithmicrequire{\textbf{Input:}}
\algrenewcommand\algorithmicensure{\textbf{Output:}}
\footnotesize
\begin{algorithmic}[1] 
\Require point\_xy\_image(XY), point\_list(PL), conic\_opacity(CO), features(F), depths(D)
\Ensure out\_alpha(OA), out\_color(OC), out\_depth(OD)
\Function{gaussianParallelRenderCUDA}{}
    \State Initialize local variables c(color), w(weight), d(depth), t, w\_t
    \State Load XY, PL, CO to share memory
    \For{p in 32 pixels} $\hspace*{\fill}$ $\triangleright$ 32 Pixel Wise Serial
        \For{j = lane\_id; j $<$ all gaussians; j += 32}
            \State BlendParallel(XY,PL,CO,F,D,c,w,d,t,j,p,w\_t) $\hspace*{\fill}$ $\triangleright$ 32 Gaussians Wise Parallel
            \State InitNext32GaussiansWithLastGaussian(w\_t)
        \EndFor
    \EndFor
    \State OA, OC, OD $\leftarrow$ WriteOutputs(c, w, d, t)
\EndFunction
\\
\Function{kernelLaunch}{}
    \State gaussianParallelRenderCUDA$<<<grid, block>>>$
\EndFunction

\end{algorithmic}
\end{algorithm}

\subsection{Fine-Grained Combined Load Balancing}
\label{sect:4.3}
\noindent The two load balancing methods among thread blocks and threads are orthogonal. We can combine them for better performance. Before that, we find that the image tiling size has a significant influence on the load balance, as well as the overall performance. Here the "tiling size" refers to the number of pixels that are assigned to a single thread block. \footnote{We define that the pixel block used in the pre-processing calculation is referred to as "patch", and the pixel block utilized by the thread block in the render calculation is named as "tile".}

\begin{figure*}[t]
  \centering
   \includegraphics[width=0.99\linewidth]{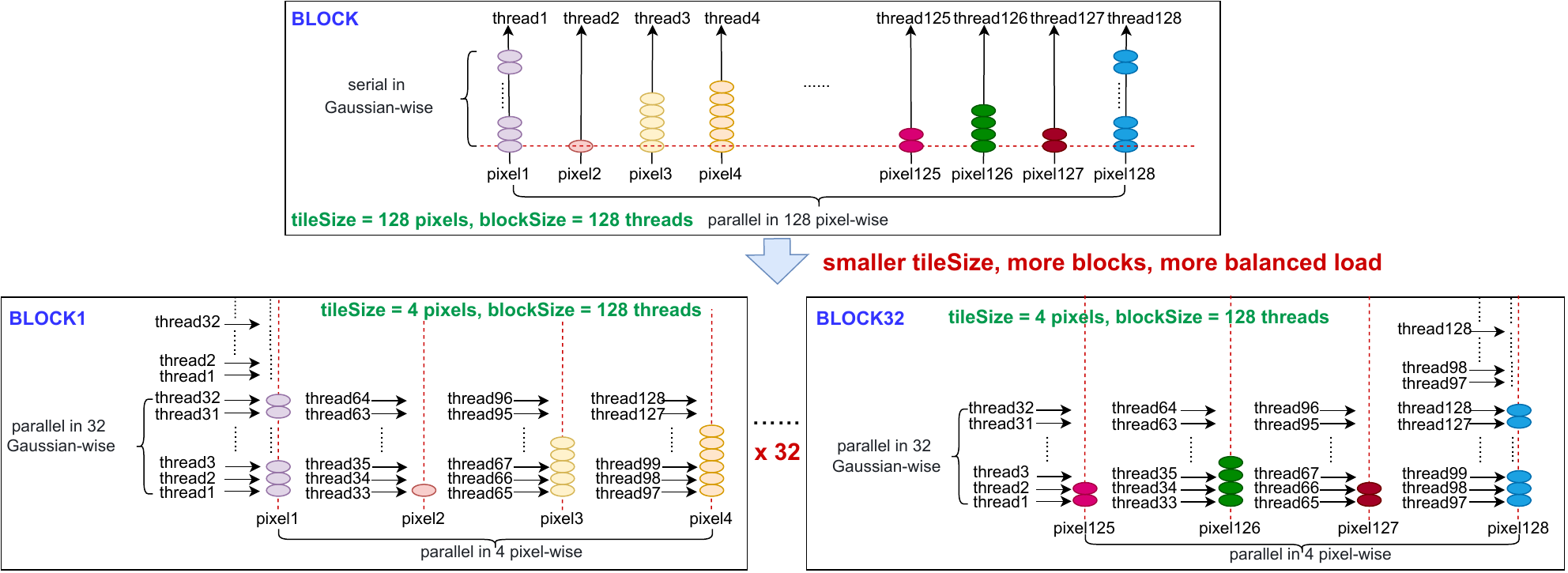}
   \caption{Fine-grained combined load balancing effect diagram.}
   \label{fig:warpsandblocks}
\end{figure*}

Here we use A100 GPU cards to demonstrate why tiling size has a significant influence on the performance. We assume that one thread block contains 128 threads for the simplicity of analysis. A100 has 108 stream multiprocessors (SMs), and each SM can schedule 64 warps (or 16 blocks with 4 warps in a block) at the same time. Therefore, one wave \footnote{A wave refers to a group of thread blocks that are executed concurrently. more waves usually mean more work distribution units, and better flexibility.} includes as many as 16*108 = 1728 thread blocks. The original method uses one thread block to deal with one tile with pixel size 16*8. Considering an image with 960*540 pixels, we can launch $\lceil960/16\rceil * \lceil540/8\rceil$ = 4080 thread blocks, which means there are only 4080/1728 = 2.36 waves in the render kernel.  Load imbalance among tiles may be significant, and we need much larger waves to assure that light-workload tiles workload can accumulate to be comparable to possible extreme heavy-workload tiles. 

To solve this, we use a smaller tile size assigned to one thread block(Algorithm \ref{alg:fine-grained-load-balancing-render}). Specifically, the patch size before rendering kernel and thread size inside a thread block both remain 16*8, but we use one thread block to deal with a 4-pixel pattern instead of a 128-pixel patch. This will bring 32x more assignable task groups(Algorithm \ref{alg:fine-grained-load-balancing-render} line 23), therefore, the load balance among blocks is feasible. In the naive render kernel, the Gaussians of a 128-size patch only need be loaded into shared memory of a block once; since now we use a block to deal with 4 pixels, those Gaussians will be loaded into shared memory by corresponding thread blocks many more times(Algorithm \ref{alg:fine-grained-load-balancing-render} line 13). The tile size of workload distribution is obviously a trade-off between load balance and memory load efficiency. But the Fig. \ref{fig:gaussiandistribution} (Gaussians distribution) and our experiments show that addressing load imbalance will bring much more performance benefits.  

After using a smaller task distribution size for blocks, we can directly combine the two load-balancing methods introduced in Section \ref{sect:4.1} and Section \ref{sect:4.2}. Fig. \ref{fig:warpsandblocks} shows our combined load balancing methods.

\begin{algorithm}
\caption{Fine-Grained Load Balancing Render Computations}
\label{alg:fine-grained-load-balancing-render}
\algrenewcommand\algorithmicrequire{\textbf{Input:}}
\algrenewcommand\algorithmicensure{\textbf{Output:}}
\footnotesize
\begin{algorithmic}[1] 
\Require point\_xy\_image(XY), point\_list(PL), conic\_opacity(CO), features(F), depths(D)
\Ensure out\_alpha(OA), out\_color(OC), out\_depth(OD)
\Function{fineGrainedLoadBalancingRender}{total\_tiles, next\_tile}
    \While {true}
        \State Synchronize block
        \If {thread\_idx $==$ 0}
            \State my\_tile\_id $\leftarrow$ atomicAdd(next\_tile, 1)
        \EndIf
        \State Synchronize block
        \If {my\_tile\_id $>$ total\_tiles}
            \State return
        \EndIf
        \State Map warp 0, 1, 2, 3 to pixel 0, 1, 2, 3 respectively
        \State Initialize local variables c(color), w(weight), d(depth), t, w\_t
        \State Load XY, PL, CO to share memory
        \For{j = lane\_id; j $<$ all gaussians; j += 32}
            \State BlendParallel(XY,PL,CO,F,D,c,w,d,t,j,w\_t) $\hspace*{\fill}$ $\triangleright$ 32 Gaussians Wise Parallel
            \State InitNext32GaussiansWithLastGaussian(w\_t)
        \EndFor
        \State OA, OC, OD $\leftarrow$ WriteOutputs(c, w, d, t)
    \EndWhile
\EndFunction
\\
\Function{kernelLaunch}{}
    \State total\_tiles $\leftarrow$ grid.x $\times$ grid.y $\times$ grid.z $\times$ 32, each block deals with 4 pixels
    \State next\_tile $\leftarrow$ 0
    \State max\_hw\_resource $\leftarrow$ 108 (A100 SM) $\times$ 16 (16 blocks with 4 warps in a block)
    \State fineGrainedLoadBalancingRender$<<<max\_hw\_resource, block>>>$(total\_tiles, next\_tile)
\EndFunction

\end{algorithmic}
\end{algorithm}

\subsection{Self-Adaptive Kernel Selection Strategy}
\label{sect:4.4}
\begin{figure}[htbp]
  \centering
   \includegraphics[width=0.99\linewidth]{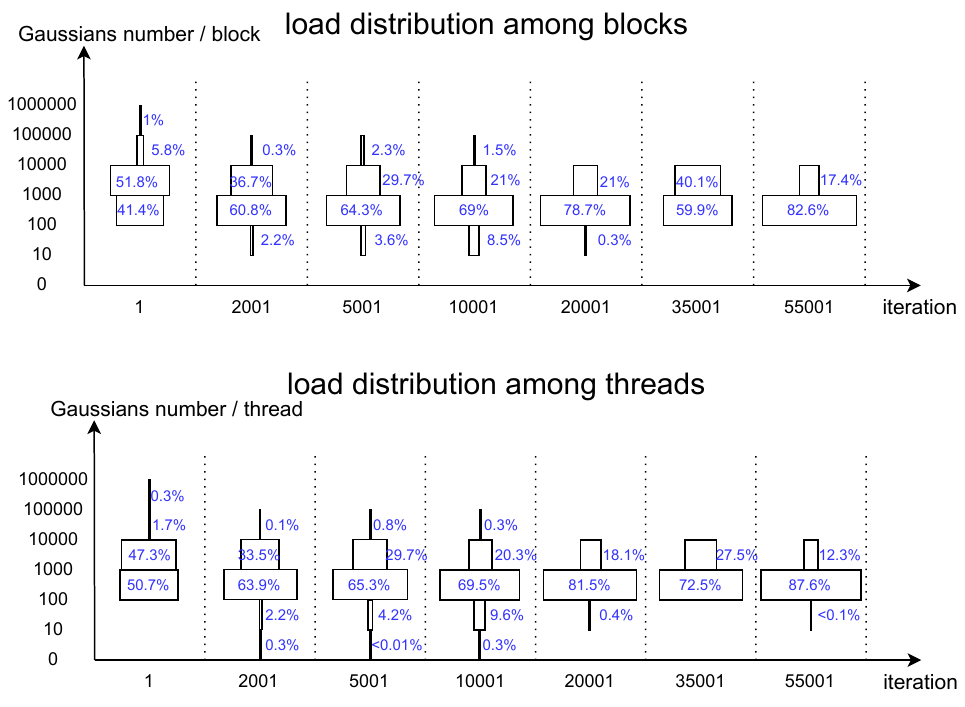}
   \caption{Gaussians load distribution among blocks and threads at different training iterations.}
   \label{fig:gaussiandistribution}
\end{figure}

\noindent The initial data for 3D Gaussian Splitting model training mostly comes from processed LiDAR sparse SfM point cloud data. The trained data features in different training steps vary a lot. The Gaussians of a tile is the key metric of load balance. Fig. \ref{fig:gaussiandistribution} shows how Gaussians distribution per block/thread changes as the training goes. In the beginning of the training, the number of Gaussians on different blocks/threads has a large difference, and some extreme-scale blocks/threads, if not processed properly, may lead to significant performance drop. In this case, we should use our proposed combined load balance optimizations. However, as the training continues, the imbalance in Gaussian number gets better. As we can see from Fig. \ref{fig:gaussiandistribution}, after some iterations of training, the maximum Gaussian number decreases significantly, and their distribution becomes more concentrated. In this case, still using our load balancing optimizations will not bring much performance benefits; besides, as we introduced above, our optimizations can cause extra overhead (warp-level reduction in Section \ref{sect:4.2} and block-level redundant memory load in Section \ref{sect:4.3}). Based on those observations, we should give up those optimizations when the data is already in good balance. Naturally, how to tell whether the data is in good balance, and accordingly choosing the best kernel is key for optimal performance.

Unlike traditional neural network training, a fixed render kernel cannot guarantee balanced workloads in 3DGS training process, so an experiment-based self-adaptive kernel selection strategy is needed, which can avoid manually choosing possible metrics for measuring the data balance, and better reflect the ground truth.

\begin{figure}[htbp]
  \centering
   \includegraphics[width=0.99\linewidth]{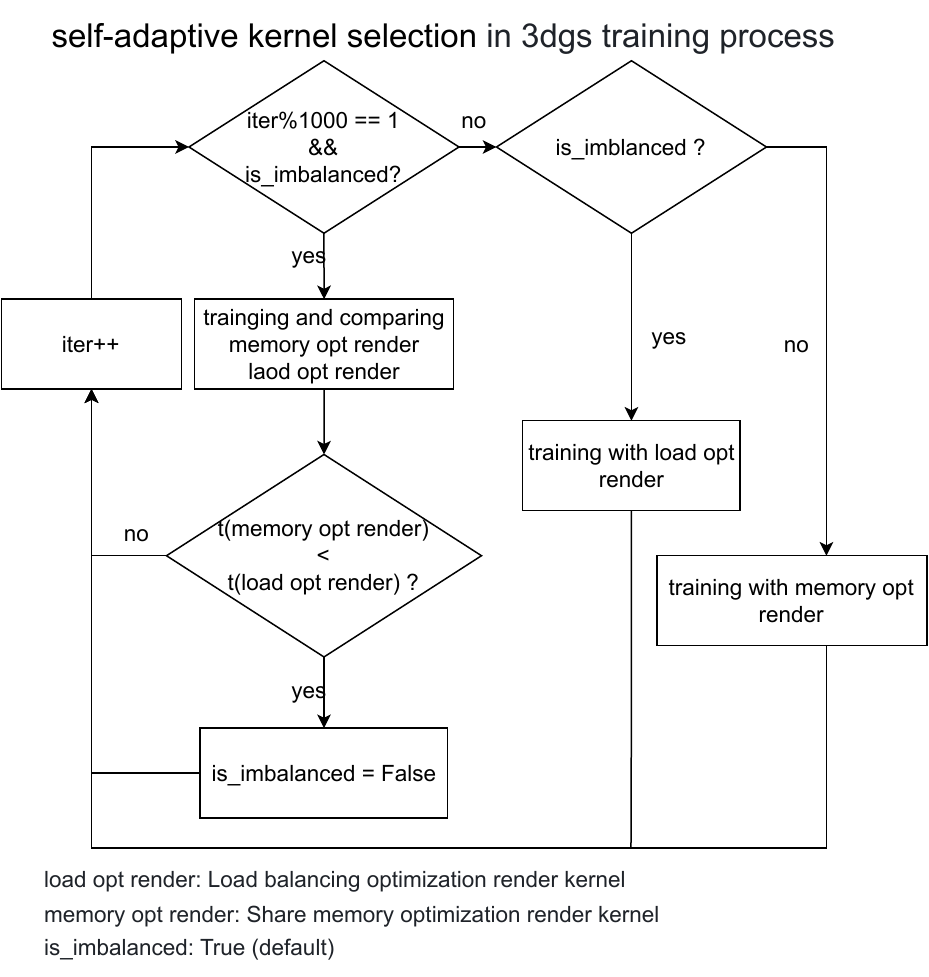}
   \caption{Schematic diagram of self-adaptive kernel selection strategy}
   \label{fig:diagram}
\end{figure}

Fig. \ref{fig:diagram} demonstrates our training process. We start with our workload-balancing kernels. Every 1000 iterations, we compare the performance of two kernels by running two kernels with current data separately. If the workload-balancing kernel has worse performance, we assume that the data is already in good balance, so  the remaining training will use the original kernel. For the original kernel, we also make some memory-access optimizations, as can be seen in Section \ref{sect:5.2.2}.

\section{Experiments}
\subsection{Experimental Setup}
\noindent In this experiment, we utilize NVIDIA’s Nsight series performance analysis tools, specifically Nsight Compute\cite{NsightCompute}, which focuses on analyzing the execution efficiency of specific CUDA kernels, such as the optimized renderCUDA kernel discussed in this paper.
The hardware platform is based on the NVIDIA A100 SMX4 80GB, and the software environment utilizes PyTorch 1.13 with CUDA version 11.8. The dataset is derived from the internal 3D Gaussian model training smoke test dataset of Li Auto. The rasterization module during the 3D Gaussian training process is the main focus of the work, which is primarily based on the open-source project Street Gaussians\cite{yan2024street}. The specific experiment mainly aims to evaluate the execution time of the forward renderCUDA kernel in the rasterization module under different optimization methods.

\subsection{Experiment Processes and Result analysis}
\subsubsection{Load Balancing Optimization Experiment}
To better evaluate the forward render kernel, we selected the data from the first iteration of the training process as the evaluation target. The size of the 2D pixel plane is (960, 540), and the preprocessing patch size is (16, 8), we sequentially applied the optimization methods mentioned above while utilizing the Nsight Compute tool. The test results are shown in Table \ref{tab:experiments}.\\\\
\textbf{Naive Experiment}. First, we conduct a naive experiment. Based on the CUDA static distribution method, image pixel size and patch size, the kernel launch configuration can be easily calculated, that is, $grid = (\lceil960/16\rceil, \lceil540/8\rceil, 1)$, $block = (16, 8, 1)$(Algorithm \ref{alg:naive render} line 11).
\begin{algorithm}
\caption{Naive Render Computations}
\label{alg:naive render}
\algrenewcommand\algorithmicrequire{\textbf{Input:}}
\algrenewcommand\algorithmicensure{\textbf{Output:}}
\footnotesize
\begin{algorithmic}[1] 
\Require point\_xy\_image(XY), point\_list(PL), conic\_opacity(CO), features(F), depths(D)
\Ensure out\_alpha(OA), out\_color(OC), out\_depth(OD)
\Function{renderCUDA}{}
    \State Initialize local variables c(color), w(weight), d(depth), t
    \State Load XY, PL, CO to share memory
    \For {j in all gaussians}
        \State BlendInOrder(XY,PL,CO,F,D,c,w,d,t,j) $\hspace*{\fill}$ $\triangleright$ Pixel Wise Parallel
    \EndFor
    \State OA, OC, OD $\leftarrow$ WriteOutputs(c, w, d, t)
\EndFunction
\\
\Function{kernelLaunch}{}
    \State renderCUDA$<<<grid, block>>>$
\EndFunction
\end{algorithmic}
\end{algorithm}

As can be seen from Table \ref{tab:experiments} (a), the time of naive render kernel is 125.5ms, and only $22.6\%$ occupancy is achieved, which is far from the upper limit of theoretical occupancy, mainly due to the load imbalance at CUDA level.\\\\
\textbf{Inter-Block Dynamic Workload Distribution Experiment}. In the case of dynamic loading distribution, the kernel launch configuration is calculated based on hardware compute resources(Algorithm \ref{alg:inter-block-load-balancing} line 23). To obtain the current processing tile\_id, the total workload quantity total\_tiles and the current processing tile\_id need to be passed as parameters to the kernel(Algorithm \ref{alg:inter-block-load-balancing} line 1, 21, 22, 24). Additionally, a loop for workload distribution logic must be included within the kernel function to achieve dynamic load balancing scheduling(Algorithm \ref{alg:inter-block-load-balancing} line 2).

As shown in Table \ref{tab:experiments} (b), the kernel execution time has been reduced to 120ms, only a $4.6\%$ improvement compared to the naive case, and the achieved occupancy is just $19.82\%$, showing a significant gap compared to the theoretical occupancy limit of $75\%$. So the load imbalance problem still exists. As explained in 4.3, the dynamic distribution capability is not strong enough to comprehensively address the load imbalance issues if the tile size(16, 8) remains unchanged.\\\\
\textbf{Gaussian-Wise Parallel Rendering Experiment}. The number of Gaussians processed by different CUDA threads varies from hundreds to hundreds of thousands shown in Fig. \ref{fig:gaussiandistribution}. Thus We conduct warp-collaboration experiment, which replaces Pixel-wise parallelism with Gaussian-wise parallelism to reduce workload divergence inside a warp. The kernel launch configuration is the same as naive experiment(Algorithm \ref{alg:gaussian-wise-parallel-render}).

As shown in \ref{tab:experiments} (c), the kernel execution time has decreased to 107.36ms, representing a $16.8\%$ performance improvement compared to the naive one. However, there is still a significant gap between achieved occupancy and theoretical occupancy, and the load imbalance issues persist.\\\\
\textbf{Fine-Grained Combined Load Balancing Experiment}. To further eliminate the load imbalance issues, we combine inter-block dynamic workload distribution and Gaussian-wise parallel rendering approaches with fine-grained pixel tile. Specifically, the size of patch and thread block remains (16, 8) (Algorithm \ref{alg:fine-grained-load-balancing-render} line 26), but we use one thread block to deal with a 4-pixel tile instead of a 128-pixel tile(Algorithm \ref{alg:fine-grained-load-balancing-render} line 23). Thus the total task tiles are increased by 32x and each warp focuses on one pixel(Algorithm \ref{alg:fine-grained-load-balancing-render} line 11 and line 23). Although the memory access overhead will also increase by 32x(Algorithm \ref{alg:fine-grained-load-balancing-render} line 13), it accounts for a small proportion in severe load imbalance scenarios. In this way, we can comprehensively address the load imbalance issues among warps and blocks.

As shown in \ref{tab:experiments} (d), the kernel execution time has been reduced to 16.69ms, 7.52X faster than the naive version. Additionally, the achieved occupancy is now very close to the theoretical occupancy, indicating the load imbalance issues among warps and blocks have been completely solved. This time, the achieved occupancy is mainly limited by the number of required registers, which is highly related to algorithm logic.\\
\begin{table*}[htbp]
\setlength\tabcolsep{3pt}
\caption{Experiments Comparison}
\centering
\scalebox{0.80}{
\begin{tabular}{lcccc} 
\toprule 
\textbf{Experiments} & \textbf{Time(ms)} & \textbf{Theoretical Occupancy(\%)} & \textbf{Achieved Occupancy(\%)} & \textbf{Occupancy Limiters} \\
\midrule 
\textbf{(a) Native} & 125.50 & 100.00 & 22.60 & Load Imbalance \\
\textbf{(b) Inter-Block Dynamic Workload Distribution} & $120.00\ (1.05x \uparrow)$ & 75.00 & 19.82 & Load Imbalance \\
\textbf{(c) Gaussian-Wise Parallel Rendering} & $107.36\ (1.17x \uparrow)$ & 31.25 & 18.85 & Load Imbalance \\
\textbf{(d) Fine-Grained Combined Load Balancing} & $\textbf{16.69}\ (\textbf{7.52x} \uparrow)$ & 62.50 & 55.02 & \textbf{Used Registers Number} \\
\bottomrule 
\label{tab:experiments}
\end{tabular}
}
\end{table*}

\subsubsection{Self-Adaptive Kernel Selection Experiment}
\label{sect:5.2.2}
\noindent As 3DGS training progresses, the training data transitions from an initially unbalanced distribution to a more balanced state as shown in Fig. \ref{fig:gaussiandistribution}. Based on the naive render kernel, we also make some memory access optimization. Specifically, we put features and depths to share memory too (refer to Algorithm \ref{alg:naive render} line 3). Although this optimization has little effect in load imbalance situations at  training initial stage due to the relatively small proportion of memory access overhead, with the training process going, the effect of shared-memory access optimization is increasingly evident owing to the progressively balanced load.

\begin{figure}[htbp]
  \centering
   \includegraphics[width=0.95\linewidth]{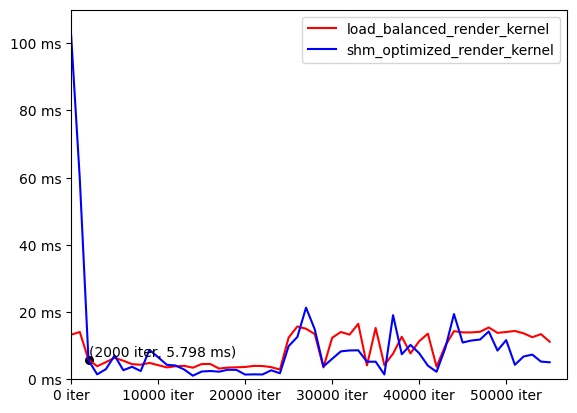}
   \caption{Optimized render kernel performance during 3DGS training process.}
   \label{fig:3DGSRenderKernelPerf}
\end{figure}

Next, we collect performance data during the training process using both combined load-balancing render kernel and shared-memory optimized render kernel, as shown in Fig. \ref{fig:3DGSRenderKernelPerf}. It can be observed that from 0 to 2000 training steps, the combined load-balancing kernel performs much better than the shared-memory optimized kernel. After 2000 steps, the combined load-balancing kernel exhibits a slightly inferior performance. The primary reason for this decline is that as the training process iterates, the distribution of Gaussians approaches a balanced state, in which scenario fine-grained combined load balancing optimization causes extra overhead, namely warp-level reduction and block-level redundant memory load described in Section \ref{sect:4.4}. It’s important to note that, in a load balanced state, the memory access overhead is non-negligible. Additionally, the increased usage of registers also limits the upper bound of warp occupancy. These factors together result in worse performance for the combined load-balancing kernel.

Given that the data workload distribution during the 3D-Gaussian model training process transitions from severe imbalance in the initial stages to a more balanced state later on, shown in Fig. \ref{fig:gaussiandistribution}. We propose an experiment-based self-adaptive kernel selection strategy during the training process to accelerate training efficiency. Specifically, during the initial stage of the model training, we choose fine-grained combined load balancing render kernel by default. Then, after every 1000 (which can be configured before training start) iterations, we compare the performance of these two kernels(fine-grained combined load balancing render kernel and the shared-memory optimized render kernel), until a performance inflection point is reached, which occurs when the distribution of Gaussians becomes balanced. After this point, subsequent iterations will select the shared-memory optimized render kernel.

\begin{figure*}[htbp]
    \begin{subfigure}[b]{0.52\textwidth}
        \includegraphics[width=1.1\textwidth]{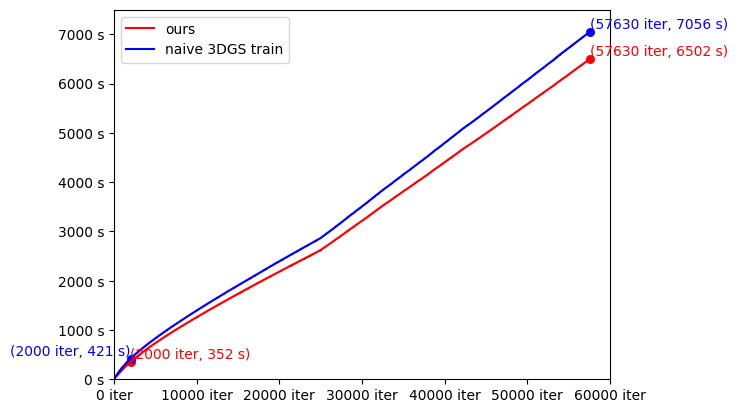}
        \caption{Left}
        \label{fig:a}
    \end{subfigure}
    \hfill
    \begin{subfigure}[b]{0.52\textwidth}    
        \includegraphics[width=\textwidth]{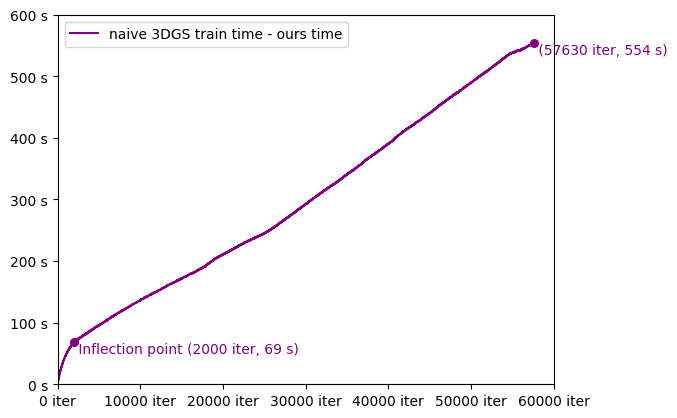}
        \caption{Right}
        \label{fig:b}    
    \end{subfigure} 
    \caption{(a) Left: 3DGS e2e training time (naive vs ours); (b) Right: 3DGS e2e training time Difference(naive - ours)}
    \label{fig:3DGS-training-difference}
\end{figure*}

The e2e performances of 3DGS training with self-adaptive kernel selection strategy and naive 3DGS training are shown in Fig. \ref{fig:3DGS-training-difference}. It can be easily obtained that from the beginning of training to the inflection point, compared to the naive version, the optimized training speeds up by $19.6\%$; From the inflection point to the end of model training, the optimized training speeds up by $7.9\%$. Overall, the optimized training speeds up by $8.5\%$.

\subsection{Discussion and Limitations}
\noindent Through our experiments, we have completely resolved the problem of poor rendering performance attributed to load imbalance during 3DGS training initial stage, and the more extreme the load imbalance scenarios are, the better the performance will be. Although our current method has also shown certain effectiveness in 3DGS multiple GPUs training, a systematic analysis from the perspective of 3DGS multiple GPUs training has not been conducted yet. In the next step, we will combine dynamic load balancing technology of 3DGS multiple GPUs distributed training\cite{scalingup} with our approaches to further enhance 3DGS training efficiency in large-model or large-scale scene reconstruction scenarios.

\section{Conclusion}
\label{sec:conclusion}
\noindent We introduce Balanced 3DGS, a Gaussian-wise parallelism rendering with fine-grained tiling approach in 3DGS training process, perfectly solving load-imbalance issue. In load imbalance scenarios, the forward renderCUDA kernel performance can be significantly enhanced through inter-block dynamic workload distribution, Gaussian-wise parallel rendering and fine-grained combined load balancing techniques. At the same time, 3DGS can self-adaptively choose the best renderCUDA kernel during the training process based on different load-balance situations, which effectively improves training efficiency.

{\small
\bibliographystyle{ieee_fullname}
\bibliography{egbib}
}

\appendix
\section{Implementation Details}
\noindent Algorithm \ref{alg:detail naive render} and \ref{alg:detail gaussian-wise parallel render} show the difference of two parallelism methods in warp-level computation. Here we concentrate on the difference of warp-level thread divergency and collaboration, and omit computation details of ${power}$, ${alpha}$ and ${t}$ in Algorithm \ref{alg:detail naive render} and \ref{alg:detail gaussian-wise parallel render}.

In pixel-parallelism kernel, one thread gets a pixel (Algorithm \ref{alg:detail naive render} line 2), and iterates over all concerned gaussian spheres (Algorithm \ref{alg:detail naive render} line 5). For each pixel and each gaussian sphere: after some related computation (Algorithm \ref{alg:detail naive render} line 7), if it's a valid gaussian sphere (Algorithm \ref{alg:detail naive render} line 8) and the computation of current pixel has not finished (Algorithm \ref{alg:detail naive render} line 12), the thread will update this pixel with related variables (Algorithm \ref{alg:detail naive render} line 17). Besides, $t$ is a variable in ${pixel}$ we should pay attention to, because each update of t depends on the results of calculations of all previous gaussian spheres. When we implement gaussian sphere dimension parallelism, we should also ensure the correctness of $t$.

In our proposed kernel, we use gaussian sphere dimension 32-way parallelism, and pixels are processed in serial (Algorithm \ref{alg:detail gaussian-wise parallel render} line 4). We use warp-level primitives to assure the warp works as a whole in determining execution paths (Algorithm \ref{alg:detail gaussian-wise parallel render} line 13 and line 23). To assure the correct accumulation of $t$, we use warp shuffle primitives (Algorithm \ref{alg:detail gaussian-wise parallel render} line 12-17) to multiply all $alpha$ in every precursor threads (Algorithm \ref{alg:detail gaussian-wise parallel render} line 15), and then broadcast the $t$ in last thread to all threads as the $t$ for next gaussian sphere group (Algorithm \ref{alg:detail gaussian-wise parallel render} line 23). Besides, each $pixel$ in each thread now is a partially updated results, so we need to warp-level reduction (Algorithm \ref{alg:detail gaussian-wise parallel render} line 28) to obtain the same result as that in Algorithm \ref{alg:detail naive render}.

In this way, we implement a gaussian sphere dimension parallelism kernel. Warp naturally runs in SIMT manner, therefore, we can avoid thread divergency by enabling threads inside a warp to run similar workload. Besides, the usage of warp-level primitives assure that intra warp thread communications will not introduce much overhead.

\begin{algorithm}
\caption{detail naive render computations}
\label{alg:detail naive render}
\algrenewcommand\algorithmicrequire{\textbf{Input:}}
\algrenewcommand\algorithmicensure{\textbf{Output:}}
\footnotesize
\begin{algorithmic}[1] 
\Require point\_xy\_image(XY), point\_list(PL), conic\_opacity(CO), features(F), depths(D)
\Ensure out\_alpha(OA), out\_color(OC), out\_depth(OD)
\Function{renderCUDA}{}
    \State pixel = pixels[lane\_id]
    \State Initialize local variables c(color), w(weight), d(depth), t
    \State Load XY, PL, CO to share memory
    \For{j = 0; j $<$ gaussians.size() and !finished; j ++}
        \State Initialize local variables alpha, power, tmp\_t 
        \State pixelGaussianCompute(XY, PL, CO, j, alpha, power)
        \If {power $>$ 0 or alpha $<$ 1/255}
            \State continue
        \EndIf
        \State tmp\_t = t * (1 - alpha)
        \If {tmp\_t $<$ 0.0001}
            \State finished = True
            \State break
        \EndIf
        \State t = tmp\_t
	    \State accumulateToLocalVariables(alpha,tmp\_t, F, D, c, w, d)
    \EndFor
    \State OA, OC, OD $\leftarrow$ WriteOutputs(c, w, d, t)
\EndFunction
\end{algorithmic}
\end{algorithm}

\begin{algorithm}
\caption{detail gaussian-wise parallel render computations}
\label{alg:detail gaussian-wise parallel render}
\algrenewcommand\algorithmicrequire{\textbf{Input:}}
\algrenewcommand\algorithmicensure{\textbf{Output:}}
\footnotesize
\begin{algorithmic}[1] 
\Require point\_xy\_image(XY), point\_list(PL), conic\_opacity(CO), features(F), depths(D)
\Ensure out\_alpha(OA), out\_color(OC), out\_depth(OD)
\Function{gaussianParallelRender}{}
    \State Initialize local variables c(color), w(weight), d(depth), t
    \State Load XY, PL, CO to share memory
    \For {p = 0; p $<$ 32; p ++}
        \State pixel = pixels[p];
        \For{j = lane\_id; j $<$ gaussians.size() and !finished; j += 32}
            \State Initialize local variables alpha, power, tmp\_t, one\_alpha, w\_c, w\_w, w\_d
	        \State pixelGaussianCompute(XY, PL, CO, j, alpha, one\_alpha, power)
            \If {\_\_all\_sync(power $>$ 0 or alpha $<$ 1/255)}
			     \State continue
            \EndIf
		    \For {offset = 1; offset $<$ 32; offset *= 2} 
			     \State v = \_\_shfl\_up\_sync(one\_alpha, offset, 32)
			     \If {lane\_id \% 32 $\ge$ offset}
				    \State one\_alpha *= v
                 \EndIf
		    \EndFor
		    \State tmp\_t = t * one\_alpha
		    \If {\_\_any\_sync(tmp\_t $<$ 0.0001)} 
			     \State finished = True
		    \EndIf
                \State t = tmp\_t
                \State t = \_\_shfl\_sync(t, 31)
		    \If {valid\_gassian\_spheres}
			     \State accumulateToLocalVariables(alpha, tmp\_t, F, D, w\_c, w\_w, w\_d)
            \EndIf
        \EndFor
        \State c, w, d $\leftarrow$ warp\_reduce(w\_c, w\_w, w\_d)
    \EndFor
    \State OA, OC, OD $\leftarrow$ WriteOutputs(c, w, d, t)
\EndFunction
\end{algorithmic}
\end{algorithm}

\end{document}